\documentclass[10pt,twocolumn,letterpaper]{article}

\usepackage[pagenumbers]{cvpr} 

\usepackage{graphicx}
\usepackage{amsmath}
\usepackage{amssymb}
\usepackage{booktabs}

\usepackage{multicol}
\usepackage{multirow}

\usepackage{pifont}
\newcommand{\cmark}{\ding{51}}%
\newcommand{\xmark}{\ding{55}}%

\usepackage[pagebackref,breaklinks,colorlinks]{hyperref}

\usepackage[capitalize]{cleveref}
\crefname{section}{Sec.}{Secs.}
\Crefname{section}{Section}{Sections}
\Crefname{table}{Table}{Tables}
\crefname{table}{Tab.}{Tabs.}

\def\cvprPaperID{9010} 
\def\confName{CVPR}
\def\confYear{2023}

\begin{document}

\title{CVT-SLR: Contrastive Visual-Textual Transformation for Sign Language Recognition with Variational Alignment}

\author{
\textbf{Jiangbin Zheng\textsuperscript{1}, Yile Wang\textsuperscript{1,2}, Cheng Tan\textsuperscript{1}, Siyuan Li\textsuperscript{1}}, \\ \textbf{Ge Wang\textsuperscript{1}, Jun Xia\textsuperscript{1}, Yidong Chen\textsuperscript{3}, Stan Z. Li\textsuperscript{1}\thanks{Corresponding author.}}\\
\textsuperscript{1}AI Lab, Research Center for Industries of the Future, Westlake University\\
\textsuperscript{2}Institute for AI Industry Research (AIR), Tsinghua University\\
\textsuperscript{3}School of Informatics, Xiamen University\\
{\tt\small \{zhengjiangbin,wangyile,tancheng,lisiyuan,wangge,xiajun,Stan.ZQ.Li\}@westlake.edu.cn}\\
{\tt\small ydchen@xmu.edu.cn}\\
}
\maketitle


\begin{abstract}
Sign language recognition (SLR) is a weakly supervised task that annotates sign videos as textual glosses. Recent studies show that insufficient training caused by the lack of large-scale available sign datasets becomes the main bottleneck for SLR. Most SLR works thereby adopt pretrained visual modules and develop two mainstream solutions. The multi-stream architectures extend multi-cue visual features, yielding the current SOTA performances but requiring complex designs and might introduce potential noise. Alternatively, the advanced single-cue SLR frameworks using explicit cross-modal alignment between visual and textual modalities are simple and effective, potentially competitive with the multi-cue framework. In this work, we propose a novel contrastive visual-textual transformation for SLR, CVT-SLR, to fully explore the pretrained knowledge of both the visual and language modalities. Based on the single-cue cross-modal alignment framework, we propose a variational autoencoder (VAE) for pretrained contextual knowledge while introducing the complete pretrained language module. The VAE implicitly aligns visual and textual modalities while benefiting from pretrained contextual knowledge as the traditional contextual module. Meanwhile, a contrastive cross-modal alignment algorithm is designed to explicitly enhance the consistency constraints. Extensive experiments on public datasets (PHOENIX-2014 and PHOENIX-2014T) demonstrate that our proposed CVT-SLR consistently outperforms existing single-cue methods and even outperforms SOTA multi-cue methods.
The source codes and models are available at \url{https://github.com/binbinjiang/CVT-SLR}.
\end{abstract}

\begin{figure*}[t]
  \centering
    \includegraphics[width=0.92\linewidth]{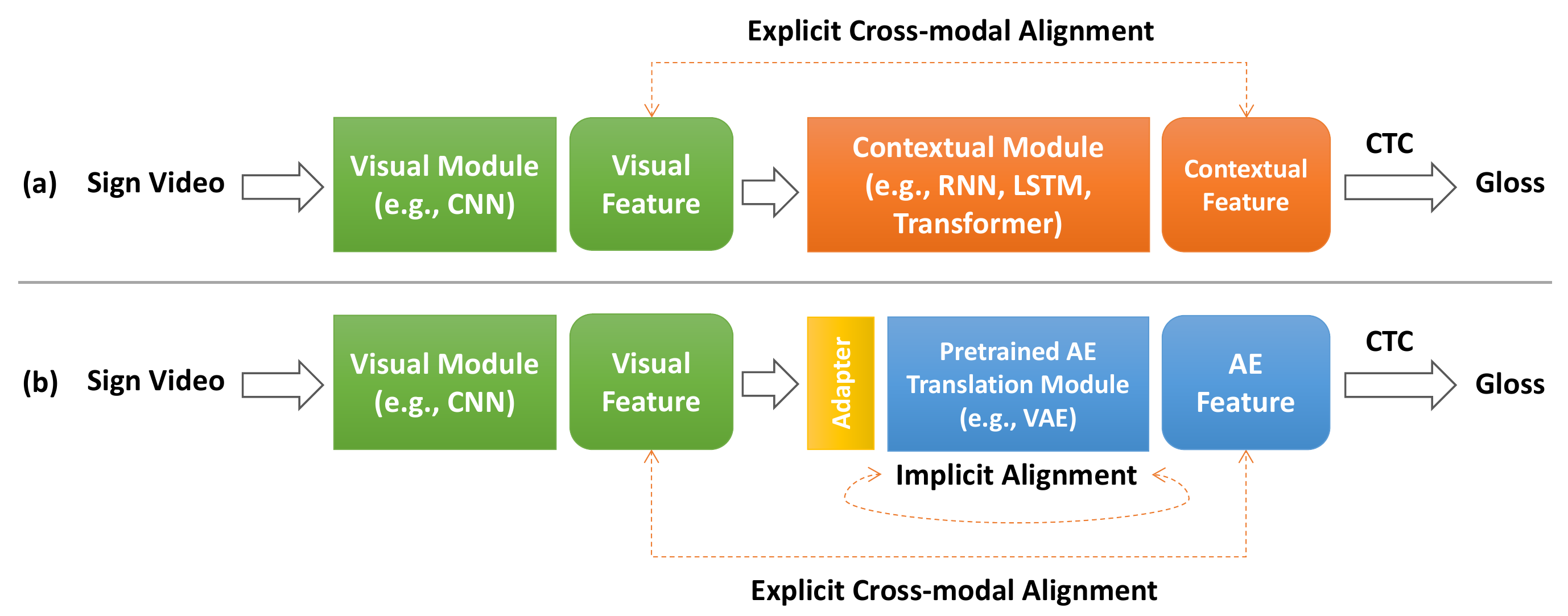}
    \vspace{-0.7em}
    \caption{(a) An advanced single-cue SLR framework with explicit cross-modal alignment; (b) Our proposed single-cue SLR framework with explicit cross-modal alignment and implicit autoencoder alignment. Both frameworks use pretrained visual features. But our framework uses the autoencoder module to replace the mainstream contextual module, which not only includes the functions of the contextual module but also can introduce complete pretrained language knowledge and implicit cross-modal alignment. To maximize the preservation of the complete pretrained language parameters and migrated visual features, a video-gloss adapter is introduced.}
   \label{fig:1}
\end{figure*}

\section{Introduction}
\label{sec:intro}

As a special visual natural language, sign language is the primary communication medium of the deaf community \cite{kamal2019technical}. With the progress of deep learning \cite{he2016deep,wu2020fuzzy,tong2020document,bao2021beit,li2022efficient}, sign language recognition (SLR) has emerged as a multi-modal task that aims to annotate sign videos into textual sign glosses. However, a significant dilemma of SLR is the lack of publicly available sign language datasets. For example, the most commonly-used PHOENIX-2014 \cite{koller2015continuous} and PHOENIX-2014T \cite{camgoz2018neural} datasets only include about 10K pairs of sign videos and gloss annotations, which are far from training a robust SLR system with full supervision as typical vision-language cross-modal tasks \cite{radford2021learning}. Therefore, data limitation that may easily lead to insufficient training or overfitting problems is the main bottleneck of SLR tasks.

The development of weakly supervised SLR has witnessed most of the improvement efforts focus on the visual module (e.g., CNN) \cite{cihan2017subunets,cui2017recurrent,pu2019iterative,niu2020stochastic,pu2020boosting,hao2021self}. Transferring pretrained visual networks from general domains of human actions becomes a consensus to alleviate the low-resource limitation. The mainstream multi-stream SLR framework extends the pretrained visual module with multi-cue visual information \cite{koller2017re,koller2019weakly,camgoz2020sign,zhou2020spatial,zheng2021enhancing,zuo2022c2slr}, including global features and regional features such as hands and faces in independent streams. The theoretical support for this approach comes from sign language linguistics, where sign language utilizes multiple complementary channels (e.g., hand shapes, facial expressions) to convey information \cite{camgoz2020sign}. The multi-cue mechanism essentially exploits hard attention to key information, yielding the current SOTA performances. However, the multi-cue framework is complex (e.g., cropping multiple regions, requiring more parameters), and the fusion of multiple streams might introduce additional potential noise.

Another mainstream advanced solution is the single-cue cross-modal alignment framework \cite{min2021visual,hao2021self}, which consists of a pretrained visual module followed by a contextual module (e.g., RNN, LSTM, Transformer) and a Connectionist Temporal Classification (CTC)  \cite{graves2006connectionist} based alignment module for gloss generation, as shown in Figure \ref{fig:1} (a). Explicit cross-modal alignment constraints further improve feature interactions\cite{min2021visual,hao2021self,tan2021co}, which could be treated as a kind of consistency between two different modalities \cite{zuo2022c2slr} facilitating the visual module learn long-term temporal information from contextual module\cite{gao2022simvp,tan2022temporal}. The cross-modal alignment framework is simple and effective, potentially competitive with the multi-cue framework.
Despite the advanced performance of complex multi-cue architectures with pretrained visual modules, the cross-modal consistency is a more elegant design for practical usage. It also implies the potential of prior contextual linguistic knowledge, which has been overlooked by existing SLR works.


In this work, we propose a novel contrastive visual-textual transformation framework for SLR, called CVT-SLR, to fully explore the pretrained knowledge of both the visual and language modalities, as shown in Figure \ref{fig:1} (b). Based on the single-cue cross-modal alignment framework, CVT-SLR keeps the pretrained visual module but replaces the traditional contextual module with a variational autoencoder (VAE). Since a full encoder-decoder architecture is used, the VAE is responsible for learning pretrained contextual knowledge based on a pseudo-translation task while introducing the complete pretrained language module. In addition, the VAE maintains the consistency of input and output modalities due to the form of an autoencoder, playing an implicit cross-modal alignment role. Furthermore, inspired by contrastive learning \cite{chen2020simple,chen2020big,caron2020unsupervised,radford2021learning}, we introduce a contrastive alignment algorithm that focuses on both positive and negative samples to enhance explicit cross-modal consistency constraints. 
Extensive quantitative experiments conducted on the public datasets PHOENIX-2014 and PHOENIX-2014T demonstrate the advance of the proposed CVT-SLR framework. Through ablation study and qualitative analysis, we also verify the effectiveness of introducing pretrained language knowledge and the new consistency constraint mechanism.

Our main contributions are as follows:
\vspace{-0.5em}
\begin{itemize}
\item A novel visual-textual transformation-based SLR framework is proposed, which introduces fully pretrained language knowledge for the first time and provides new approaches for other cross-modal tasks. 
\vspace{-0.5em}
\item New alignment methods are proposed for cross-modal consistency constraints: a) exploiting the special properties of the autoencoder to implicitly align visual and textual modalities; b) introducing an explicit contrastive cross-modal alignment method.
\vspace{-0.5em}
\item The proposed single-cue SLR framework not only outperforms existing single-cue baselines by a large margin but even surpasses SOTA multi-cue baselines.

\end{itemize}

\section{Related Work}
\label{sec:relatedwork}

SLR tasks are generally divided into three categories \cite{kamal2019technical}: finger-spelling recognition \cite{peng2009chinese,ji20163d,pan2018sign}, isolated word recognition \cite{wang2015fast,wang2016isolated,zhuang2017towards,liang20183d} and continuous sign sentence recognition \cite{cui2019deep,pu2019iterative,de2019spatial,niu2020stochastic,cheng2020fully}. In the early days, SLR works mainly focused on lexical-level tasks, such as finger-spelling recognition and isolated word recognition. Nowadays, the more practical continuous sign sentence recognition task has become mainstream sign language research. In this work, the mentioned SLRs refer specifically to continuous sign sentence recognition.

The recent SLR works \cite{cui2019deep,cheng2020fully,niu2020stochastic,hao2021self} are summarized based on these three aspects: feature extraction, recognition, and alignment. Mostly, the components of feature extraction are composed of the visual module and the contextual module. The visual module encodes short-term spatial information, while the contextual module encodes long-term context information. Based on the extracted features, the classifiers can get a posterior probability for recognition. And the alignment module is required to find the proper alignment between clips and glosses to ensure an accurate training procedure.

A few studies extend the common SLR framework by incorporating multi-cue information \cite{koller2017re,koller2019weakly,zhou2020spatial,zheng2022leveraging,zheng2021enhancing}. The multi-cue features mainly include information from hand shapes, facial expressions, mouths, and poses. The current SOTA SLR works are based on multi-cue mechanisms, such as $\text{C}^2\text{SLR}$ \cite{zuo2022c2slr}. Recently, some works show that explicitly enforcing the consistency between the visual and textual modules can also get comparable performance, although only single-cue features are considered. For example, VAC \cite{min2021visual} treats the visual and textual modules as the student and teacher, respectively, and achieves knowledge distillation to align the visual and the textual modalities. Similarly, SMKD \cite{hao2021self} achieves knowledge transfer by further sharing classifiers between different modalities. Note that SMKD is the current SOTA single-cue method.

It is worth mentioning that neural sign language translation \cite{camgoz2018neural,zheng2020improved,camgoz2020sign,chen2022simple} is another popular sign language task. As a distinction, the neural sign language translation task aims to convert sign videos into equivalent spoken language translations, further focusing on semantic relationships and adjusting word order \cite{camgoz2018neural}, while the SLR task only annotates sign videos into gloss sequences in the frame order.

\begin{figure*}[t]
  \centering
   \includegraphics[width=0.9\linewidth]{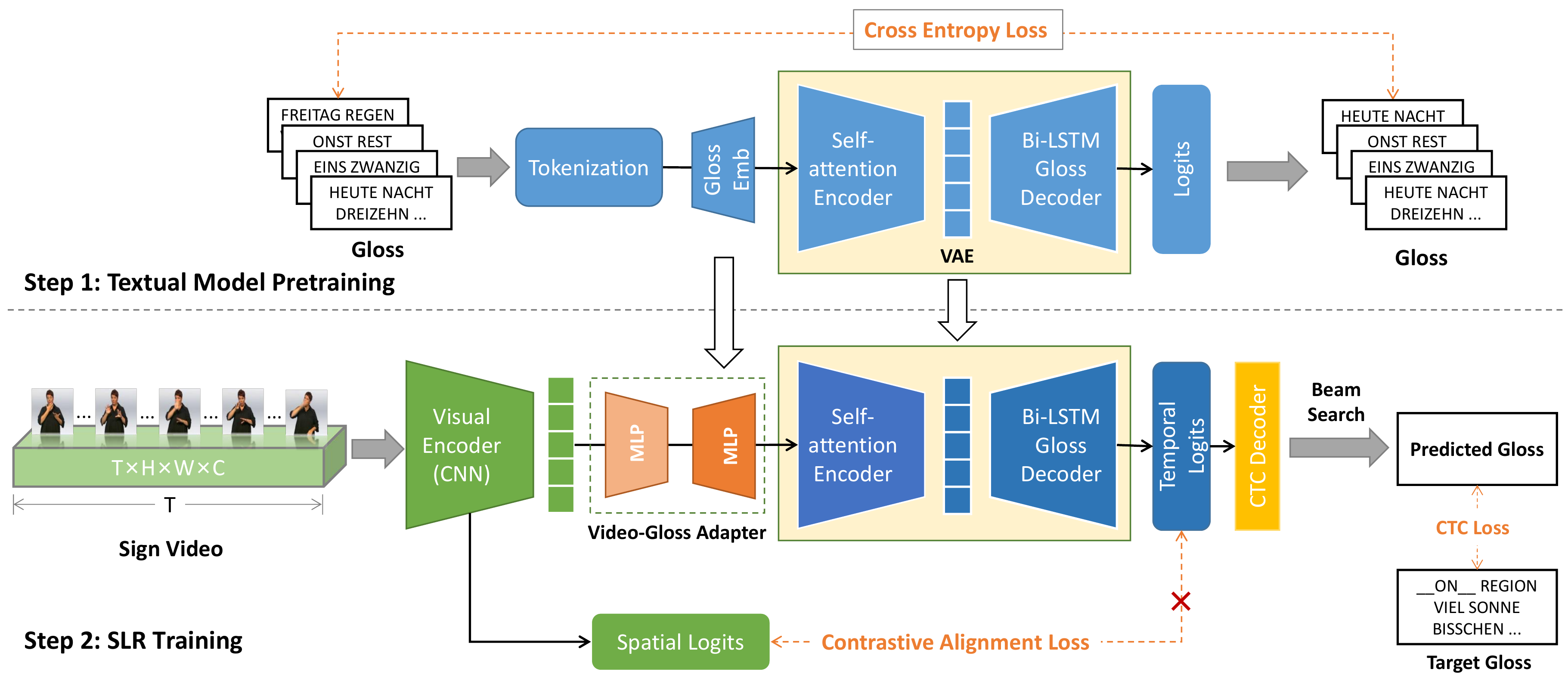}
   \vspace{-0.5em}
   \caption{Overview of proposed CVT-SLR framework. Our training pipeline is divided into two steps. Step 1: Pretrain the VAE module with the self-supervised gloss-to-gloss pseudo-translation task. Step 2: Train the CVT-SLR based on the VAE model from Step 1.}
   \vspace{-0.5em}
   \label{fig:2}
\end{figure*}

\section{Methods}
\label{sec:methods}
As shown in Figure \ref{fig:2}, our training pipeline is roughly divided into two steps. Step 1 is to pretrain a VAE network as the textual module. And step 2 is to transfer the off-the-shelf visual module (usually a publicly available CNN trained on Kinetics/ImageNet) and the pretrained textual module from step 1 into the CVT-SLR framework. A video-gloss adapter is used to bridge these two pretrained modules. 

\subsection{Textual Model Pretraining}
Our VAE model is an asymmetric encoder-decoder architecture, where the encoder consists of self-attention layers, and the decoder consists of Bi-LSTM layers, as shown in the top part of Figure \ref{fig:2}. To pretrain the VAE for our proposed SLR framework solely based on the gloss data from the PHOENIX-14 training set (excluding the development set and testing set), we construct a pseudo-translation task, Gloss2Gloss, which enables the VAE to learn the contextual semantic knowledge. Formally, it is assumed that the input gloss of the encoder is denoted as $G$, and the generative gloss of the decoder is denoted as $G^*$.

\textbf{Posterior Self-attention Encoder}. We employ the neural approximation, and reparameterization \cite{kingma2013auto,rezende2014stochastic} techniques to approximate the posterior self-attention encoder for generating the posterior distribution $q_{\phi}(z|G)$, where $\phi$ denotes the variational parameters of the encoder. The $q_{\phi}(z|G)$ is a diagonal Gaussian distribution $N(\mu, \text{diag}(\sigma^{2}))$, and two single-layer MLPs are used to parameterize the mean $\mu$ and the variance $\sigma^2$, respectively, as:
\begin{equation}
z = \mu + \sigma \odot \epsilon,
\label{eq:3-1}
\end{equation}
where $\epsilon$ is a standard Gaussian noise, and $\odot$ denotes an element-wise product.

\textbf{Generative Bi-LSTM Decoder}.
Given the continuous latent variable $z$, then the translation target $G^{*}$ is generated from $z$ as:
\begin{equation}
  p_{\theta}(G^{*},z) = p_{\theta}(G^{*}|z)p_{\theta}(z),
\end{equation}
where $\theta$ denotes the parameters of the generative Bi-LSTM decoder, and $p_{\theta}(G^{*}|z)$ is the conditional distribution that models the generation procedure estimated via the decoder.

\textbf{Variational Optimization}. Both the generative model $p_{\theta}(G^{*},z)$ and the posterior inference model $q_{\phi}(z|G)$ are incorporated tightly in an end-to-end manner, and we can apply a standard back-propagation to optimize the gradient of the variational lower bound:
\begin{equation}
\begin{split}
\mathcal{L}_{\text{VAE}}(\theta,\phi;G) = -\text{KL}(q_{\phi}(z|G)||p_{\theta}(z)) \\
+\mathbb{E}_{q_{\phi}(z|G)} {[\text{log} p_{\theta}(G^{*}|z)] \leq \text{log} p_{\theta}(G^{*})},
\end{split}
\label{eq:3-3}
\end{equation}
where $\text{KL}$ denotes the Kullback-Leibler divergence and the prior $p_{\theta}(z)$ denotes a standard Gaussian distribution. Practically, the first $\text{KL}$ term of Eq.~(\ref{eq:3-3}) can be written as:
\begin{equation}
   \mathcal{L}_{\text{KL}} = - \frac{1}{2} (\text{log} \sigma^{2} - \sigma^{2}-\mu^{2} + 1).
\end{equation}

Since our main target is a Gloss2Gloss translation task, we employ a cross-entropy loss to maximize the generative gloss probability as the second term of Eq.~(\ref{eq:3-3}):
\begin{equation}
  \mathcal{L}_{\text{gloss2gloss}} = \text{CrossEntropy}(G, G^{*}).
\end{equation}

Therefore, the overall optimization objective of textual model pretraining is a joint loss of $\mathcal{L}_{\text{KL}}$ and $\mathcal{L}_{\text{gloss2gloss}}$ as:
\begin{equation}
   \mathcal{L}_{\text{VAE}} = \mathcal{L}_{\text{KL}} + \mathcal{L}_{\text{gloss2gloss}} .
\end{equation}


\subsection{Training CVT-SLR Framework}
\subsubsection{Training Pipeline}
In the CVT-SLR framework, we sequentially concatenate the pretrained visual CNN module and the pretrained textual VAE module through a video-gloss adapter, as shown in Figure \ref{fig:2}. The pretrained ResNet18 model \cite{he2016deep} is selected as the frame-wise feature extractor, which is trained on large-scale human action datasets. For the VAE module, although Gaussian noise is introduced in the reparameterization process in step 1, we do not add any noise after migrating the VAE to the SLR framework in step 2. In this way, the reparameterization module of the VAE degenerates into two common linear layers, i.e., $\epsilon$ is removed in Eq.~(\ref{eq:3-1}).

Formally, given a sign video with $T$ RGB frames $\mathcal{F} = \{\mathcal{F}_t\}^T_{t=1} \in \mathbb{R}^{T \times H \times W \times 3}$, the visual module, which is composed of a stack of 2D-CNN layers and a global average pooling layer, first extracts spatial features $v = \{v_t\}^T_{t=1} \in \mathbb{R}^{T \times d}$. Then the textual module further extracts temporal features $s = \{s_t\}^T_{t=1} \in \mathbb{R}^{T \times d}$ . Finally, the alignment module utilizes CTC to compute the conditional probability of the gloss label sequence $p(\mathcal{G}|\mathcal{F})$ based on temporal features $s$, where $\mathcal{G} = \{\mathcal{G}_i\}^N_{i=1}$ is a generated sign gloss sequence with $N$ glosses. A shared classifier is used between the visual module and the textual module to generate spatial and temporal probability, respectively. And the proposed contrastive cross-modal alignment algorithm further enhances the consistency across different modalities.

\subsubsection{Video-Gloss Adapter}
To maximize the preservation of both pretrained visual and textual knowledge, we use a video-gloss adapter to connect spatial-temporal features, as shown in Figure \ref{fig:2}. The video-gloss adapter is simply implemented as a fully-connected MLP with two hidden layers. We take the CNN output probabilities as the input of the adapter where the final fully connected layer is initialized using the weights of the pretrained word embedding from the pretrained VAE \cite{chen2022simple}.

Since both the input and output are in the same textual form during the textual model pretraining in step 1, the VAE plays an implicit cross-modal alignment role after being transferred to the CVT-SLR framework in step 2, which is close to the effect of consistency. Therefore, the video-gloss adapter is essentially a mediator for activating the consistency between the spatial and temporal modalities.

\subsubsection{CTC Alignment}
Due to the spatial-temporal nature of sign videos, glosses have a one-to-many mapping to video frames but share the same ordering. Unlike spoken language texts, the continuous gloss annotations are chronologically consistent with the sign videos. Concretely, for an input sign video $\mathcal{F}$ and the corresponding generated gloss sequence $\mathcal{G}$, we use CTC to compute $p(\mathcal{G}|\mathcal{F})$ by marginalizing over all possible $\mathcal{F}$ to $\mathcal{G}$ alignments:
\begin{equation}
p(\mathcal{G}|\mathcal{F}) = \sum_{\pi \in \mathcal{B}} p(\pi|\mathcal{F}),
\end{equation}
where $\pi$ denotes a path and $\mathcal{B}$ is the set of all viable paths corresponding to $\mathcal{G}$. The probability $p(\pi|\mathcal{F})$ is computed by the textual module. The CTC loss is then formulated as:
\begin{equation}
\mathcal{L}_{\text{ctc}} = - \text{log } p(\mathcal{G}^{*}|\mathcal{F}),
\end{equation}
where $\mathcal{G}^{*}$ is the ground truth gloss sequence.


\begin{table*}[htp]
\centering
\scalebox{0.98}{
\begin{tabular}{clccccl} \toprule

\multirow{2}{*}{\textbf{Groups}} & \multirow{2}{*}{\textbf{Models}} & \multicolumn{2}{c}{\textbf{Dev (\%)}} & \multicolumn{2}{c}{\textbf{Test (\%)}} & \multirow{2}{*}{\textbf{Cues}} \\ 

\cmidrule(lr){3-4}\cmidrule(lr){5-6}

    &  & \textbf{DEL/INS} & \textbf{WER} & \textbf{DEL/INS} & \textbf{WER} & \\ \midrule

\multirow{13}{*}{Group 1}	
    & SubUNet \cite{cihan2017subunets}	& 14.6/4.0	& 40.8	& 14.3/4.0	& 40.7 & video \\
	& Staged-Opt \cite{cihan2017subunets}		& 13.7/7.3	& 39.4	& 12.2/7.5	& 38.7 & video \\
	& Align-iOpt \cite{pu2019iterative} & 12.6/2.6	& 37.1	& 13.0/2.5	& 36.7 & video \\
	& DPD+TEM \cite{zhou2019dynamic}	& 9.5/3.2	& 35.6  & 9.3/3.1 & 34.5 & video \\
	& Re-Sign \cite{koller2017re}	& -	& 27.1	& -	& 26.8 & video \\
	& SFL \cite{niu2020stochastic}	& 7.9/6.5	& 26.2	& 7.5/6.3	& 26.8 & video \\
	& DNF \cite{cui2019deep}	& 7.8/3.5	& 23.8 & 7.8/3.4	& 24.4 & video \\
	& FCN \cite{cheng2020fully}	& -	& 23.7	& -	& 23.9 & video \\
	& VAC \cite{min2021visual} & 7.9/2.5	& 21.2	& 8.4/2.6	& 22.3 & video \\
	& CMA \cite{pu2020boosting}	& 7.3/2.7	& 21.3	& 7.3/2.4	& 21.9 & video \\
	& SFL \cite{niu2020stochastic} & 10.3/4.1	& 24.9	& 10.4/3.6 & 25.3 & video \\
    & VL-SLT \cite{chen2022simple} & -	& 21.9	& - & 22.5 & video \\
	& SMKD	\cite{hao2021self} & 6.8/2.5	& \underline{20.8} & 6.3/2.3	& \underline{21.0} & video \\
 \midrule
 
\multirow{3}{*}{Group 2}
    & DNF \cite{cui2019deep}	& 7.3/3.3	& 23.1	& 6.7/3.3	& 22.9 & video+optical flow \\
	& STMC \cite{zhou2020spatial}	& 7.7/3.4	& 21.1	& 7.4/2.6	& 20.7 & video+hand+face+pose \\
	& $\text{C}^2\text{SLR}$	\cite{zuo2022c2slr}	&  - & \underline{20.5}	& - & \underline{20.4} & video+keypoints \\
	 \midrule

\multirow{4}{*}{Group 3}
    & $\text{Ours}_{1}$ (\textit{w/o} \text{VAE+Contra})  & 7.1/3.0	& 21.1	& 7.3/2.9 & 21.4 & video \\
	& $\text{Ours}_{2}$ (\textit{w/} \text{VAE})	& 6.5/2.4 & 20.2 & 6.3/2.2	& 20.3 & video \\
	& $\text{Ours}_{3}$ (\textit{w/} \text{Contra}) &  6.7/2.7 & 20.4	& 6.4/2.5 & 20.7 & video \\
	& $\text{Ours}_{4}$ (\textit{w/} \text{VAE+Contra}) &  6.4/2.6 & \textbf{19.8}	& 6.1/2.3 & \textbf{20.1} & video \\
	 \bottomrule
\end{tabular}
}
\vspace{-0.25em}
\caption{\label{table:1} Performance comparison (\%) on PHOENIX-14 dataset. DEL/INS: deletion error and insertion error. The best results and SOTA baseline for each group are marked as \textbf{bold} and \underline{underlined}.}
\vspace{-0.3em}
\end{table*}

\subsubsection{Contrastive Alignment}
Aligning both visual and textual modalities help improve SLR \cite{min2021visual}. To enforce alignment constraints across modalities, we propose a contrastive alignment loss. In addition, shared classifiers are also used as an improvement technique between visual and textual features \cite{hao2021self}.

In previous SLR works, cross-modal alignment only focuses on positive samples\cite{zhou2020spatial,hao2021self}. Inspired by contrastive learning \cite{chen2020simple,he2020momentum,zheng2022using}, we construct both positive and negative samples in the same mini-batch and implement a contrastive cross-modal alignment method to ensure that similar features are closer while different features are farther apart. Given that the normalized spatial features from CNN as $\mathcal{S}_{\text{logits}} \in \mathbb{R}^{B \times T \times d}$, and the normalized temporal features from VAE as $\mathcal{V}_{\text{logits}} \in \mathbb{R}^{B \times T \times d}$,  where $B$ denotes the number of samples. Then we compute pair matrices firstly as:
\begin{equation}
\mathcal{S}2\mathcal{V}_{\text{pair}} =\mathcal{S}_{\text{logits}} \times \mathcal{V}_{\text{logits}}^T,
\end{equation}
\begin{equation}
\mathcal{V}2\mathcal{S}_{\text{pair}} =\mathcal{V}_{\text{logits}} \times \mathcal{S}_{\text{logits}}^T ,
\end{equation}

Taking $\mathcal{S}2\mathcal{V}_{\text{pair}} \in \mathbb{R}^{B \times B}$ pair matrix (denoted as $\mathcal{P}$) as an example, $\mathcal{P}[i,j] (0 \leq i,j < B)$ represents the similarity value between the visual feature of the $i$-th batch and the textual feature of $j$-th batch. The visual and textual features from the same input instances are positive samples, and the features from other different input instances are negative samples. Hence, in the matrix $\mathcal{P}$, the diagonal similarity values are from positive sample pairs, and the rest values are from negative sample pairs. To make the loss computation differentiable and easy, we convert the computation of positive and negative pairs into a binary classification task\cite{radford2021learning}. The value $i$ corresponding to the $i$-th row in the matrix $\mathcal{P}$ is a positive label. Therefore, the corresponding labels of $\mathcal{P}$ are $\text{Labels}=\{0, 1, \cdots, i, \cdots, B\}$, then the contrastive alignment loss related to $\mathcal{S}2\mathcal{V}_{\text{pair}}$ as:
\begin{equation}
\hspace{-0.5em}\mathcal{L}_{\text{align}_\mathcal{S}} = \text{CrossEntropy}(\text{Softmax}(\mathcal{S}2\mathcal{V}{\text{pair}}), \text{Labels}).
\end{equation}
Similarly, the $\mathcal{V}2\mathcal{S}_{\text{pair}}$ temporal-to-spatial alignment loss as:
\begin{equation}
\mathcal{L}_{\text{align}_\mathcal{V}} = \text{CrossEntropy}(\text{Softmax}(\mathcal{V}2\mathcal{S}_{\text{pair}}), \text{Labels}).
\end{equation}
Therefore, the complete contrastive alignment loss as:
\begin{equation}
\mathcal{L}_{\text{align}} = \frac{1}{2}(\mathcal{L}_{\text{align}_\mathcal{S}} + \mathcal{L}_{\text{align}_\mathcal{V}}) .
\end{equation}


Overall, the objective of CVT-SLR is jointly optimized CTC loss $\mathcal{L}_{\text{ctc}}$ and contrastive alignment loss $\mathcal{L}_{\text{align}}$ as:
\begin{equation}
   \mathcal{L}_{\text{CVT-SLR}} = \mathcal{L}_{\text{ctc}} + \mathcal{L}_{\text{align}}.
\end{equation}

\begin{table*}[t]
\vspace{-0.4em}
\centering
\scalebox{0.95}{
\begin{tabular}{clccl}\toprule

\multirow{2}{*}{\textbf{Groups}} & \multirow{2}{*}{\textbf{Models}} & \multicolumn{2}{c}{\textbf{WER}} & \multirow{2}{*}{\textbf{Cues}} \\ 
\cmidrule(lr){3-4}

&  & \textbf{Dev(\%)} & \textbf{Test(\%)}  \\ 
\midrule

\multirow{4}{*}{Group 1}
	& SFL \cite{niu2020stochastic}	& 25.1	& 26.1 & video \\
	& CNN+LSTM+HMM \cite{koller2019weakly}	& 24.5 & 26.5 & video \\
	& SLT \cite{camgoz2020sign} & 24.9	& 24.6	& video \\
	& FCN \cite{cheng2020fully}	& 23.3	& 25.1	& video \\
	& SMKD \cite{hao2021self}	& \underline{20.8} & \underline{22.4}	& video \\
\midrule
  
\multirow{5}{*}{Group 2}
	& CNN+LSTM+HMM \cite{koller2019weakly} & 24.5	& 25.4 & video+mouth \\
	& CNN+LSTM+HMM \cite{koller2019weakly}	& 22.1 & 24.1 & video+mouth+hand \\
	& SLT \cite{camgoz2020sign}	& 24.6	& 24.5	& video+text \\
	& STMC \cite{zhou2020spatial}	& \underline{19.6}	& 21.0	& video+hand+face+pose \\
	& $\text{C}^2\text{SLR}$ \cite{zuo2022c2slr}	& 20.2 & \underline{20.4}	& video+keypoints \\
\midrule

  \multirow{4}{*}{Group 3}
    & $\text{Ours}_{1}$ (\textit{w/o} \text{VAE+Contra}) & 21.8 & 22.0	& video \\
	& $\text{Ours}_{2}$ (\textit{w/} \text{VAE}) & 20.1 & 20.4	& video \\
	& $\text{Ours}_{3}$ (\textit{w/} \text{Contra}) &  21.0 & 21.5	& video \\
	& $\text{Ours}_{4}$ (\textit{w/} \text{VAE+Contra}) & \textbf{19.4}  & \textbf{20.3}	& video \\
\bottomrule
\end{tabular}
}
\vspace{-0.5em}
\caption{\label{table:2} Performance comparison (\%) on PHOENIX14T dataset. The best results and SOTA baseline for each group are marked as \textbf{bold} and \underline{underlined}, respectively.}
\vspace{-0.8em}
\end{table*}

\section{Experiments}
\subsection{Settings}
\textbf{Datasets}. We use PHOENIX-2014 \cite{koller2015continuous} and PHOENIX-2014T \cite{camgoz2018neural} as datasets, where RGB videos and their corresponding annotations are provided. PHOENIX-2014 is a German SLR dataset about weather forecasts with a vocabulary size of 1,081, which is divided into three parts: 5,672 instances for training, 540 for validation, and 629 for testing, respectively. And PHOENIX-2014T is an extension of PHOENIX-2014 with a vocabulary size of 1,085, which is also divided into three parts: 7,096 instances for training, 519 for validation, and 642 for testing, respectively.

\textbf{Evaluation Metric}. Word error rate (WER) is the most widely-used metric to evaluate the SLR performance, which measures the number of necessary insertions, substitutions, and deletions in the recognized sentences when compared to the reference sentences. WER is an edit distance, indicating the least number of operations of substitutions ($\#\text{sub}$), insertions ($\#\text{ins}$), and deletions ($\#\text{del}$) to transform the predicted sentences into the reference sequences ($\#\text{reference}$):
\begin{equation}
\vspace{-0.3em}
\text{WER} = \frac{\#\text{del}+\#\text{ins}+\#\text{sub}}{\#\text{reference}}.
\end{equation}

\textbf{Implementation Details}. The videos are resized to $256 \times 256$ and then cropped to $224 \times 224$. During SLR training, we use random crop and horizontal flips (50\%) for data augmentation. During SLR testing, we only adopt center cropping. Beam Search is used for gloss generation. Our visual module has the same configuration as \cite{hao2021self}. The VAE module contains two multi-head self-attention layers with 4 heads and two Bi-LSTM layers with $2 \times 512$ dimensional hidden states sequentially. Adam \cite{kingma2014adam} optimizer is used to pretrain our VAE network with a batch size of 16, an initial learning rate of 1e-4. And AdamW \cite{loshchilov2018fixing} optimizer is used to train our SLR framework with a batch size of 8, a learning rate of 1e-4, and a weight decay of 1e-4. And 1 NVIDIA A100 80GB GPU is used.

\subsection{Main Results}

Table \ref{table:1} and Table \ref{table:2} show our comparisons with two-group different baselines, where Group 1 includes single-cue baselines and Group 2 includes multi-cue baselines. $\text{Ours}_{number}$ denote our trained CVT-SLR framework, which have only two variables, ``VAE'' and ``Contra''. Note that ``\textit{w/o} (\textit{w/)} VAE'' denotes the VAE module in CVT-SLR framework without (with) transferred pretrained parameters, and ``\textit{w/o} (\textit{w/)} Contra'' denotes training CVT-SLR framework without (with) contrastive alignment loss.

\textbf{Evaluation on PHOENIX-14}. Table \ref{table:1} shows our methods compared to other methods on PHOENIX-14. The optimal WER scores of our single-cue CVT-SLR ($\text{Ours}_{4}$) on the development set and testing set are 19.8\% and 20.1\%, respectively. And $\text{Ours}_{4}$ achieves the best performance among video-based only/single-cue methods in Group 1 ($\text{Ours}_{4}$ vs. $\text{SMKD}$). Furthermore, while no extra cues are used, our CVT-SLR achieves the best result among the models trained with extra cues (i.e., multi-cue methods) in Group 2 ($\text{Ours}_{4}$ vs. $\text{C}^2\text{SLR}$).

From within-group comparisons in Group 3, we find that the case, $\text{Ours}_{1}$, which neither migrates the pretrained VAE parameters nor uses the contrastive alignment constraint, performs the worst. Nevertheless, relative to other groups, it still performs well. Based on $\text{Ours}_{1}$, by only configuring the pretrained VAE parameters ($\text{Ours}_{2}$) or the contrastive alignment constraints ($\text{Ours}_{3}$), the CVT-SLR framework can also gain gains, and the pretrained VAE brings more obvious gains (+0.9\%/+1.1\% vs. +0.7\%/+0.7\%). Overall, Group 3 demonstrates that both the prior language knowledge from the pretrained textual module and the contrastive constraints through cross-modal alignment play important enhancement roles for our SLR framework.

\textbf{Evaluation on PHOENIX-14T}. In Table \ref{table:2}, we evaluate the CVT-SLR on PHOENIX-14T. We can see that our optimal model ($\text{Ours}_{4}$) in this dataset also achieves the best performance (development set: 19.4\% and testing set: 20.3\%) with video-based only/single-cue information compared to Group 1 ($\text{Ours}_{4}$ vs. $\text{SMKD}$). It is worth mentioning that the SOTA single-cue method ($\text{SMKD}$) fails to surpass the multi-cue baselines in their work on PHOENIX-14T. And our CVT-SLR makes up for their shortcomings while outperforming the SOTA multi-cue baselines in Group 2 on this dataset ($\text{Ours}_{4}$ vs. $\text{C}^2\text{SLR}$).

Compared within Group 3 on the PHOENIX-14T, the observed  conclusions are more or less the same as on the PHOENIX-14, i.e., both the pretrained textual module and the proposed contrastive alignment constraint have a beneficial effect on our CVT-SLR framework. But the pretrained textual module gain on PHOENIX-14T is more significant than that on PHOENIX-14 (+1.7\%/+1.6\% vs. +0.9\%/+1.1\%), indicating that prior language knowledge is more beneficial for this dataset.

\begin{table}
\vspace{-0.25em}
\centering
\scalebox{0.95}{
\begin{tabular}{ccccc}\toprule

\multirow{2}{*}{\textbf{\#}} & \multirow{2}{*}{\textbf{Pre-Visual}} & \multirow{2}{*}{\textbf{Pre-Textual}} & \multicolumn{2}{c}{\textbf{WER}} \\ 
\cmidrule(lr){4-5}
&  &  & \textbf{Dev(\%)} & \textbf{Test(\%)}  \\ 
\midrule
	1 & \xmark & \xmark & 89.2	& 88.5 \\
	2 & \xmark & \cmark & 30.8  & 31.4 \\
	3 & \cmark & \xmark	& 20.4 	&  20.7 \\
	4 & \cmark & \cmark	& \textbf{19.8} & \textbf{20.1} \\
\bottomrule
\end{tabular}
}
\vspace{-0.3em}
\caption{\label{table:3} Ablation study of pretrained modules on PHOENIX-14 (Pre-Visual: pretrained visual module, Pre-Language: pretrained textual module). The best results are \textbf{bolded}.}
\vspace{-0.3em}
\end{table}

\begin{table}
\centering
\scalebox{0.9}{
\begin{tabular}{ccccc}\toprule

 \multirow{2}{*}{\textbf{\#}} & \multirow{2}{*}{\textbf{CTC Weight}} & \multirow{2}{*}{\textbf{Align Weight}} & \multicolumn{2}{c}{\textbf{WER}} \\ 
\cmidrule(lr){4-5}
&  &  & \textbf{Dev(\%)} & \textbf{Test(\%)}  \\ 
\midrule
	1 & 1.0	& 0 & 20.2 & 20.3 \\
	2 & 1.0 & 1.0  &  20.0 & 20.2 \\
	3 & 1.0 & 10.0 &  \textbf{19.8}	&  \textbf{20.1} \\
	4 & 1.0 & 100.0 & 20.1  &  20.6 \\
	5 & 1.0 & 1000.0	&  20.5 & 20.9 \\
\bottomrule
\end{tabular}
}
\vspace{-0.3em}
\caption{\label{table:4} Ablation study of varied weights of the CTC loss and the alignment loss on PHOENIX-14. The best results are \textbf{bolded}.}
\vspace{-0.8em}
\end{table}

\subsection{Ablation Study}
We study the effectiveness of the proposed pretrained modules and alignment loss weights in our framework.

\begin{figure}[t]
   \centering
   \vspace{-0.5em}
   \includegraphics[width=0.95\linewidth]{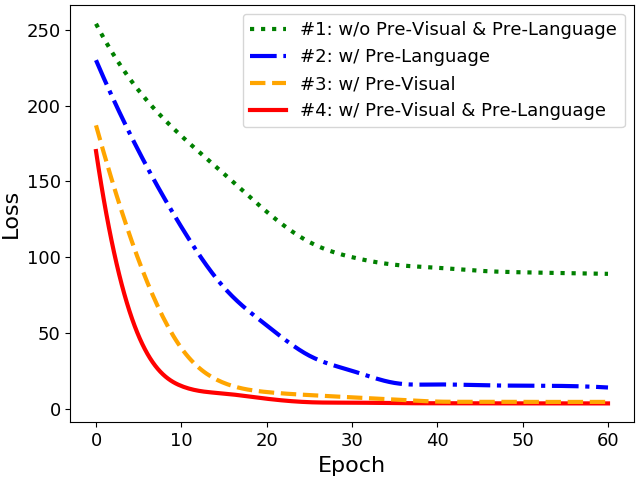}
   \vspace{-0.6em}
   \caption{The training loss curves corresponding to the ablation study of pretrained modules in Table \ref{table:3}.}
   \vspace{-0.8em}
   \label{fig:3}
\end{figure}

\begin{figure*}[t]
  \centering
  \vspace{-0.4em}
   \includegraphics[width=0.99\linewidth]{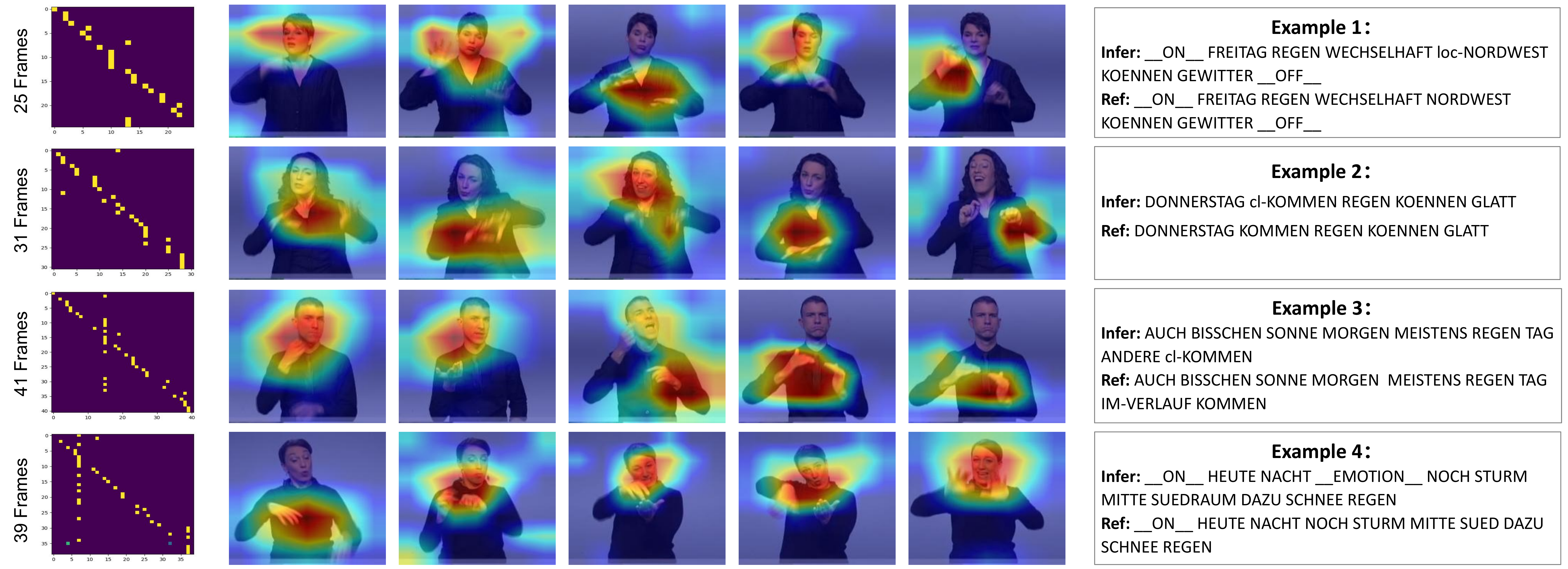}
   \vspace{-0.4em}
   \caption{Four examples with cross-modal alignment matrices (left), saliency maps (middle), and generated glosses (right) on the PHOENIX-14 test set. Cross-modal alignment matrices show the alignment between visual and textual features, while saliency maps show the highest activation regions focused on gestures.
   }
   \vspace{-0.4em}
   \label{fig:4}
\end{figure*}

\subsubsection{Pretrained Modules}
To illustrate the importance of the pretrained modules in our CVT-SLR framework, we conduct the ablation studies of both the pretrained visual module (CNN module) and the pretrained textual module (VAE module) on PHOENIX-14, as verified in Table \ref{table:3}.

Overall, both pretrained modules are useful. Comparing \#1 and \#3, we find that pretrained visual features account for a larger determinant of performance, which is consistent with our reason for retaining the pretrained visual knowledge. Surprisingly, comparing \#1 and \#2, we find that without pretrained parameters of the visual module, i.e., using only the pretrained textual module, our CVT-SLR performance is also greatly improved. Notably, this is a new observation that has not been explored in previous SLR works extensively. These comparisons demonstrate that the pretrained VAE module has learned sufficient language knowledge to improve the SLR task. Comparing \#3 with \#4 further confirms this point.

To verify the effectiveness of the pretrained modules for the training convergence speed, training loss curves corresponding to Table \ref{table:3} are shown in Figure \ref{fig:3}. Specifically, comparing \#1 and \#2 (or \#3 and \#4), configuring a pretrained textual module can promote faster convergence and reach a stable state earlier. Similarly, comparing \#1 and \#3 (or \#2 and \#4), a pretrained visual module can accelerate convergence speed compared to the pretrained textual module. As can be seen from the loss curves, the degrees of convergence are positively correlated with the final evaluation performances of WER scores.

\subsubsection{Alignment Loss Weights}
Training CVT-SLR to recognize glosses is optimized jointly by CTC loss and contrastive alignment loss. To explore the effect of loss weights, we conduct an ablation study of varied weights of the CTC loss and the proposed contrastive alignment loss, as shown in Table \ref{table:4}. For simplicity, we set the weights of CTC loss to a fixed value of 1.0. In Table \ref{table:4}, \#1, without contrastive alignment loss (weight=0), performs worse than the models with contrastive alignment in an optimal configuration (\#2, \#3), which also illustrates the gain effect of the proposed contrastive alignment. As shown in \#2-\#5 in Table \ref{table:4}, we keep increasing the weight of the alignment loss from 1.0 to 1000.0. As the weight of the contrastive loss increases, the performance starts to rise, reaching a peak when the weight value is 10.0. As the weight increases and the value is greater than 10.0, the performance starts to decline instead. This ablation study indicates the important role of the proposed contrastive alignment. By balancing the weights between CTC loss and contrastive alignment loss, the model can achieve optimal performance.

\subsection{Qualitative Analysis}
We randomly select four examples from the test set of PHOENIX-14 for qualitative analysis. Each example displays an alignment matrix, a series of saliency maps, and generated glosses in Figure \ref{fig:4}. An alignment matrix shows alignment relationships between visual and textual features, while a saliency map shows the highest activations.

\textbf{Visualization on Cross-modal Alignment Matrices}. From these four alignment matrices in Figure \ref{fig:4}, we can see that the reference and inferred glosses are almost the same with quite low WER scores, and the performances are relatively perfect. Hence, the highlighted areas of the alignment matrices are concentrated near the diagonal, which means that our contrastive alignment algorithm plays a crucial role in aligning visual and textual features.

\textbf{Visualization on Saliency Maps}. We visualize the key parts of the sign video frames in focus by using GradCAM \cite{selvaraju2017grad} as shown in Figure \ref{fig:4}. The results show that, for well-performing inferences, significant spatial features are mainly focused on the face and hand regions in the saliency maps. This observation is consistent with the sign linguistic knowledge \cite{kamal2019technical,camgoz2020sign}, where sign language conveys information mainly relying on hand shapes and facial expressions.

\section{Conclusion}
In this work, we propose a novel contrastive visual-textual transformation framework for SLR, called CVT-SLR, which further alleviates insufficient training by fully exploring the pretrained knowledge of both the visual and language modalities. We introduce prior language knowledge into the single-cue CVT-SLR for the first time via a VAE-based pretrained textual module. Furthermore, we propose new methods for cross-modal consistency constraints. One method takes advantage of the properties of the autoencoder to implicitly align the visual and textual modalities. While another method introduces a contrastive cross-modal alignment for explicit consistency. To prove the proposed CVT-SLR framework, extensive quantitative experiments and qualitative analysis are conducted, showing that our single-cue CVT-SLR not only outperforms single-cue baselines by a large margin but also surpasses SOTA multi-cue methods.

Overall, we hope our CVT-SLR can inspire the SLR community to mine pretrained modules from new aspects and design a more efficient SLR framework for practical usage. In future work, we plan to introduce a large-scale pretrained language model to guide textual module learning and further explore the potential of prior contextual linguistic knowledge for SLR tasks.

\section*{Acknowledgement}

We thank the anonymous reviewers for their constructive and helpful reviews. This work was supported by the National Key R\&D Program of China (Project No. 2022ZD0115100), the National Natural Science Foundation of China (Project No. U21A20427), the Competitive Research Fund (Project No. WU2022A009) from the Westlake Center for Synthetic Biology and Integrated Bioengineering.

{\small
\bibliographystyle{ieee_fullname}
\bibliography{egbib}
}

\end{document}